\documentclass{ecai}
\usepackage{graphicx}
\usepackage{latexsym}

\usepackage{times}
\usepackage{soul}
\usepackage{url}
\usepackage[hidelinks]{hyperref}
\usepackage[utf8]{inputenc}
\usepackage[small]{caption}
\usepackage{graphicx}
\usepackage{amsmath}
\usepackage{amsthm}
\usepackage{booktabs}
\usepackage{algorithm}
\usepackage{algorithmic}
\usepackage[switch]{lineno}

\usepackage{booktabs}
\usepackage{caption,color}
\usepackage{multirow}

\begin{document}

\begin{frontmatter}

\title{A semantics-driven methodology for high-quality image annotation}



\author[A]{\fnms{Fausto}~\snm{Giunchiglia}\orcid{0000-0002-5903-6150}}
\author[A]{\fnms{Mayukh}~\snm{Bagchi}\orcid{0000-0002-2946-5018}}
\author[A]{\fnms{Xiaolei}~\snm{Diao}\orcid{0000-0002-3269-8103}\thanks{Corresponding Author.}}

\address[A]{Department of Information Engineering and Computer Science (DISI), University of Trento, Italy.}
\address[]{\{fausto.giunchiglia, mayukh.bagchi, xiaolei.diao\}@unitn.it}

\begin{abstract}
Recent work in Machine Learning and Computer Vision has highlighted the presence of various types of systematic flaws inside ground truth object recognition benchmark datasets. Our basic tenet is that these flaws are rooted in the \textit{many-to-many mappings} which exist between the \textit{visual information} encoded in images and the \textit{intended semantics} of the labels annotating them. The net consequence is that the current annotation process is largely under-specified, thus leaving too much freedom to the subjective judgment of annotators.
In this paper, we propose \texttt{vTelos}, an integrated Natural Language Processing, Knowledge Representation, and Computer Vision methodology whose main goal is to make explicit the (otherwise implicit) intended annotation semantics, thus minimizing the number and role of subjective choices. A key element of \texttt{vTelos} is the exploitation of the \textit{WordNet} lexico-semantic hierarchy as the main means for providing the meaning of natural language labels and, as a consequence, for driving the annotation of images based on the objects and the visual properties they depict. The methodology is validated on images populating a subset of the \textit{ImageNet} hierarchy.
\end{abstract}

\end{frontmatter}

\section{Introduction}
\label{S1}
\begin{figure*}[!t]
 \includegraphics[width=1\linewidth]{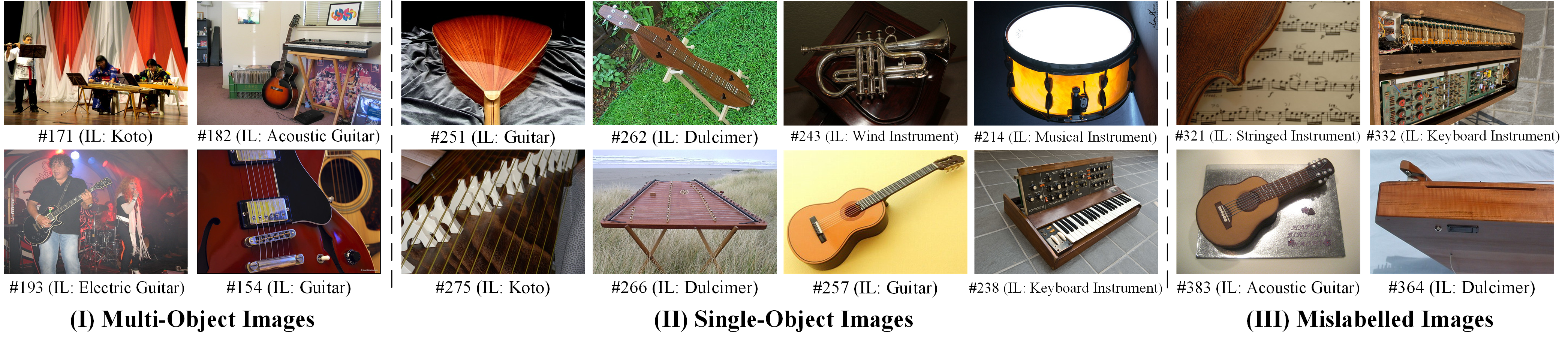}
\caption{Samples from ImageNet Musical Instrument Sub-Hierarchy  (IL stands for \textit{ImageNet Label)}.}
\centering
\label{I1}
\end{figure*}

Recent work in Machine Learning (ML) and Computer Vision (CV) has highlighted various types of systematic flaws in the development of object recognition benchmark datasets. See, for instance, \cite{2019-CVPR,2021-NeurIPS}.
Our basic tenet is that these flaws are grounded in the way perception and language interact. As a matter of fact, an early version of this problem was already identified in \cite{SGP-2000} as the \textit{Semantic Gap Problem} (SGP), which was described 
as the ``\emph{lack of coincidence between the information that one can extract from the visual data and the interpretation that the same data have for a user in a given situation}". This problem, still unsolved, was formalized in \cite{SNCS-2021} as the fact that there is a \emph{many-to-many mapping} between the \textit{visual information} encoded in an image and the \textit{intended semantics} of the corresponding linguistic descriptions. Thus, for instance, in Fig.\ref{I1}-(II), Image \#251 is labeled \textit{Guitar} but, just because it gives only a partial view of the instrument, it could have also been labeled \textit{Bass} or \textit{Ukelele}, while, at the same time, \textit{Guitar} is also the label of images \#154, \#257.
The net consequence is that the current annotation process is largely under-specified, thus leaving too much freedom to the annotator's judgment, who can then subjectively select one among the many SGP mappings.

In this paper, we propose \texttt{vTelos}\footnote{From the Greek word \texttt{Telos} which means the end; the object aimed at in an effort; \textit{purpose}, with \texttt{v} standing for \texttt{visual}, as in \textit{visual purpose}.}, an integrated Natural Language Processing (NLP), Knowledge Representation (KR) and CV methodology whose main goal is to generate high-quality datasets. There are two main underlying intuitions. The first is that the \textit{purpose} of an annotation effort, i.e., what the ML model trained by the dataset should be used for, should be made explicit via precise guidelines, that we organize around \textit{four main Design Choices}, as follows:
\begin{itemize}
\vspace{-0.1cm}
    \item C1: for each \textit{image}, which object(s) should be considered?
    \item C2: for each \textit{object}, which set of properties should be considered?
    \item C3: for each \textit{set of properties}, which label should be selected to properly describe them?
    \item C4: for each \textit{label}, if \textit{polysemic},\footnote{\label{f1} For instance, in WordNet, the average number of meanings per noun and per polysemous noun are 1.25 and 4, respectively, where the more common a noun, the higher the polysemy \cite{Freihat2014AnOA}.} how do we unambiguously represent its meaning?
    \vspace{-0.1cm}
\end{itemize} 
\noindent
The second underlying intuition is that the role of the annotator should be split in two, i.e., that of the \textit{Classificationist} and that of the \textit{Classifier}, each associated with different tasks (this terminology being adopted from KR and Knowledge Organization (KO) \cite{SRR-67}). Thus, the classificationist, will first define the space of labels to be used (Design Choices C3,C4) and, for each of them, the (visual) properties they describe. Then, in a second phase, the classifier will classify images using the labels defined by the classificationist (Design Choices C1,C2), where Choice C2 is based on the output of Choice C3. Notice how the proposed decision process is  reversed with respect to how image annotation is usually performed, as also described above.  The key idea is that we should first select the space of labels to be used, with a clear understanding of their intended meaning, in terms of the visual properties they are taken to describe, and then exploit this clarity when selecting, for each input image, the label which most properly describes its content. The proposed approach actually follows the standard practice in KR, for instance when classifying products in an online catalogue \cite{get-specific}.

In this process, \textit{the intended natural language  semantics of natural language terms} (as selected during C3, C4) \textit{and the relevant visual information of images} (i.e., objects and their visual properties, as selected during C1) \textit{are natively aligned} (during C2), this being the reason why the SGP does not appear. The annotation bias is dealt with not by building an all-encompassing dataset, which would be impossible \cite{KD-1995-bouquet,giunchiglia2021transparency} but, rather, by making explicit the annotation choices, as in \cite{erculiani2023egocentric}. 
This advantage is further amplified by the fact that, instead of performing choices C3, C4 from scratch, we take, as a predefined source of natural language labels, the \textit{WordNet Genus-Differentia lexico-semantic hierarchy} \cite{PWN}. In WordNet, each label is associated with a \textit{gloss}, which, in fact, is an \textit{unambiguous linguistic definition of the properties of the objects it describes}. This opens up the possibility of full and general alignment between natural language lexical semantics and object recognition, thus enabling a general purpose solution to the SGP many-to-many mapping problem.

There are three main contributions of this paper:

\begin{enumerate}
 \vspace{-0.1cm}
    \item A classification of the main sources of annotation mistakes;
    \item A general purpose annotation methodology, i.e., \texttt{vTelos}, for dealing with the SGP many-to-many mapping problem;
    \item The application of \texttt{vTelos} to the alignment of the linguistic semantics of labels to the visual semantics of images. Notice how multi-lingual lexical resources, see, e.g., 
  \cite{L-navigli2012babelnet},
    allows to go beyond the current English-only main practice and apply \texttt{vTelos} to any natural language.
 \vspace{-0.2cm}
\end{enumerate} 
The paper is organized as follows. Section \ref{S2} introduces the sources of the annotation mistakes. Section \ref{S3} describes the \texttt{vTelos} four design choices. Section \ref{S4} provides the main algorithm which must be followed by annotators in order to implement Choice C2, when taking in input the WordNet hierarchy. Section \ref{S5} provides an evaluation of \texttt{vTelos} on  (i) inter-annotator agreement, as a proxy of the role of subjective choices, (ii) annotation cost, and (iii) impact on quality of the ML models it generates. The evaluation is carried on a substantial subset of \textit{ImageNet} \cite{IMAGENET-2009}. The choice of ImageNet is not by chance. In fact, despite the fact that this information has hardly been used so far, ImageNet has been generated by populating WordNet, and, therefore, it natively contains the definitions encoding the meaning of labels. Furthermore, ImageNet provides a very good testbed for evaluation, having been analyzed in many studies, also focusing on annotation mistakes, see, e.g., \cite{2020-AnntPipelineError,salari2022object,northcutt2103pervasive,raji2021ai}.
 Related work (Section \ref{S6}) and conclusion (Section \ref{S7}) complete the paper.

\section{Sources of annotation mistakes}
\label{S2}
\begin{table*}[!t]
\centering
\resizebox{1\linewidth}{!}{%
\begin{tabular}{|l|l|l|l|l|}
\hline
\textit{\textbf{Role}} & \textit{\textbf{Annotation Choice}} & \textit{\textbf{Purpose}}  & \textit{\textbf{Source of Subjectivity}} & \textit{\textbf{Solution}}  \\
\hline
Classificationist & C3: Label generation & Choose label and its properties  & MI, SOII & Linguistic Genus-Differentia \\
Classificationist & C4: Label disambiguation & Choose identifier  & SOIL & Alinguistic Identifiers \\
Annotator & C1: Object localization          & Choose object  & MOI & Bounding Polygons \\
Annotator & C2: Visual classification     & Choose label via its properties & MI, SOII, SOIL, SOIA & Visual Genus-Differentia   \\
\hline
\end{tabular}}
\caption{The \texttt{vTelos} four choices. MOI and MI stand for Multi-Object and Mislabelled Images, respectively.  SOII, SOIL, SOIA stand for Single-Object Images where the source of subjectivity is the Image, the Label and the Annotator background knowledge, respectively.}
\label{T1}
\end{table*}

Let us take an \textit{annotation mistake} to be a label that does not correspond to the (classificationist) \textit{intended semantics} of the image. The sources of annotation mistakes are the cause of the SGP many-to-many mapping as they raise the possibility of choices that are different from those of the classificationist. 
Annotation mistakes have three main recurring sources, as follows (see, e.g., \cite{ICML-2020}\footnote{ \url{https://arxiv.org/abs/2005.11295} is an extended version.} for an extensive evaluation-based analysis of the problem):

\begin{enumerate}
 \vspace{-0.1cm}
\item \textit{Mislabelled Images} (MI), where the flaw arises because of mislabelling. 

\item \textit{Multi-Object Images} (MOI), where the flaw arises from the occurrence of multiple objects in the same image and the consequent possible non-correspondence between the classificationist and the classifier assigned labels.

\item \textit{Single-Object Images} (SOI), where the flaw arises from an ambiguity in how to label the single object in the image.
 \vspace{-0.1cm}
\end{enumerate}
\textit{Mislabelled images} are an extreme case of SGP many-to-many mapping, and are motivated by factors such as carelessness, missing knowledge or excessive speed, that is, all well-known crowdsourcing problems, see for instance \cite{daniel2018quality}. A concrete example is in Fig.\ref{I1}-(III), wherein Image \#383, labelled as \emph{Acoustic Guitar}, actually depicts a birthday cake. 

\vspace{0.1cm}
\noindent
\textit{Multi-Object Images} (Fig.\ref{I1}-(I)) constitute more than one-fifth of ImageNet's total images. Here the source of subjectivity is  the occurrence of multiple objects, coupled with the fact that the objects selected often do not correspond to the most likely main object. One such example is the image of a \emph{Stage} (Image \#193), having ImageNet label \emph{Electric Guitar}. Thus, on one side, empirical evidence from cognitive psychology \cite{ROSCH-1976} makes clear that the main object chosen by humans possesses the highest \emph{`cue validity'}, i.e., the one which is visually the most salient, while, on the other side, many ImageNet labels correspond to \emph{random features} which don't generalize to object recognition in the wild (see \cite{ICML-2020} for more details). 

\vspace{0.1cm}
\noindent
 \textit{Single-Object Images} (Fig.\ref{I1}-(II)), are by far the most important source of annotation mistakes. The main reason is in the underlying interaction between \emph{visual polysemy} \cite{SGP-2000} and \emph{linguistic polysemy} \cite{PWN}. Single-Object images present three types of ambiguity.

\vspace{0.1cm}
\noindent
 \textit{In the first case} (SOII), \textit{the source of subjectivity is intrinsic to the Image itself.} It occurs when an object in an image is \emph{visually polysemic}\footnote{See the definition in \url{https://gradientscience.org/benchmarks/}}, namely when its visual \emph{``semantics is described only partially"} \cite{SGP-2000} and, consequently, its linguistic description is not unique. An example is Image \#251 representing a dreadnought-shaped instrument, which is labelled \textit{Guitar} in Fig.\ref{I1}-(II), but could also be labelled, e.g., \emph{Bass} or \emph{Ukelele}. Notice that images of this type usually encode \textit{a typical partial view} with no possibility of discrimination among different objects. \cite{ICML-2020} observes that not only do humans assign an alternative label for 40\% of the images, but they also assign up to 10 different labels. 

\vspace{0.1cm}
\noindent
\textit{The second source of subjectivity} in single-object images (SOIL) \textit{is intrinsic to the meaning of Labels}. There are two cases. In the case of \textit{polysemic} labels (i.e., of one-to-many mapping between labels and images), different objects, each corresponding to one of the possible meanings, are associated with the same label\cite{giunchiglia2023incremental}. 
One such example is the two images labelled as \textit{Dulcimer} (Image \#262, \#266) in Fig.\ref{I1}-(II)) where the two objects are essentially two different musical instruments, i.e., the first \textit{Dulcimer} being native to the American and the second to the Chinese culture, and are visually very different. The study in \cite{ICML-2020} highlights how these ambiguous classes generate overlaps in the image distributions with a negative impact on performance. The second case labels which are \textit{synonyms} (i.e., of many-to-one mapping between labels and images) wherein a single set of images, will wrongly be split into two sets. One such example is that of \emph{Koto} which can also be synonymously labelled as, for instance, a \emph{Kin} or a \emph{Jusangen}\cite{2023-iConf}.

\vspace{0.1cm}
\noindent
\textit{The third source} of subjectivity in single-object images (SOIA) \textit{relates to the Annotator incomplete background knowledge}. Thus, for instance, Image \#257 is labelled as \textit{Guitar} in Fig.\ref{I1}-(II), while the image depicts an \emph{Acoustic Guitar}.
This type of generic label appears whenever correct labelling requires deep domain specific knowledge. \cite{2019-CVPR} discusses the bias which arises because of this problem in the case of minority \textit{languages}, \textit{cultures}, and also with images about \textit{low income} people. 
Notice that this is not mislabelling. The issue is that the label captures some but not all of the visual properties of the object.

Overall, the sources of annotation mistakes can be summarized by saying that MOI images generate a one-to-many mapping from images to labels, those of type SOIA generate a many-to-one mapping, while those of type MI, SOII, and SOIL generate both types of mappings. These four types of mistakes are independent of one another and can, therefore, occur together in the same image, thus generating multiple occurrences of the SGP many-to-many mapping problem.

\section{Annotation design choices}
\label{S3}
As from the introduction, the \texttt{vTelos} annotation process is articulated into four main choices C1-C4. Let us analyze them in detail following the order in which they are made.
\begin{enumerate}
\vspace{-0.1cm}
  \item \textit{Choice C3: Label generation.} Choose the space of labels used to annotate images. Here the meaning of each label is \textit{defined} in terms of linguistically defined properties encoding  a selected set of visual properties.  One such example is the definition of a \textit{guitar} as being a \textit{``string instrument with six strings"}. This eliminates any source of \textit{linguistic ambiguity} intrinsic to \textit{informally defined labels} and it allows to say that two labels are \textit{synonyms}.

    \item \textit{Choice C4: Label disambiguation.} Choose a unique concept identifier for each of the (one or more) meanings of each label. This eliminates any source of \textit{linguistic ambiguity} intrinsic to \textit{polysemous labels} (see footnote \footref{f1}).
    
    \item \textit{Choice C1: Object localization.} Choose one of the (one or more) objects in an image, thus eliminating any possible source of \textit{object ambiguity}.
    
    \item \textit{Choice C2: Visual classification.} Choose the relevant \textit{visual properties} of the object under consideration and annotate the image with one of the labels selected in Choice C3. This choice is conceptually similar to the usual annotation choice. The key difference is that what is selected is not the label but, rather, the linguistic properties that describe the visual properties of the object. Thus for instance, if the object occurring in the image is a guitar, the annotator would not choose the label \textit{Guitar} but, rather the property \textit{having six strings}. The label \textit{Guitar} would then be automatically associated to the label because of the definition of guitar generated by Choice C3. This eliminates any  source of \textit{visual ambiguity}.

\end{enumerate}
\noindent
These four choices allow us to deal with the annotation mistakes described in Section \ref{S2}. Consider Table \ref{T1}, where the first column reports the two \texttt{vTelos} roles (with \textit{Annotator} can also be read as \textit{Classifier}), the second, the design choice, the third, the purpose behind the choice, the fourth, the source of subjectivity which is addressed and the last, the solution that is enforced. 

In Table \ref{T1}, the  first and the last column are 
 crucial to the \texttt{vTelos} approach for three reasons.
The first is the splitting of the annotation tasks into two roles, i.e., \textit{classificationist} and \textit{annotator},
 where the first is in charge of preparing a well-specified annotation task for the second. The second reason is that, opposite to the annotation standard practice, in \texttt{vTelos}, Choices C3,C4 are performed before Choices C1,C2. This is in order to force the annotator to select a label within of a (possibly very large) pre-selected \textit{controlled vocabulary}. Following \texttt{vTelos} this vocabulary should be built by a domain expert. 
 The third reason is that 
 classificationist and classifier coordinate through the use of the
    \textit{Visual} and the \textit{Linguistic Genus-Differentia} (see the last column in Table \ref{I1}). The idea behind the notion of Linguistic Genus-Differentia is that the meaning of a label (e.g., \textit{Guitar}) is defined in terms of a linguistic description composed of two parts, the \textit{(linguistic) Genus}, which describes what the object has in common with other objects (in the case of a guitar the fact that it is a \textit{string instrument}), and the \textit{(linguistic) Differentia} which describes which properties differentiate the object from the other objects with the same genus (in the case of a guitar, the fact that it has \textit{six strings}). In turn, given a label whose meaning is defined in terms of Linguistic Genus-Differentia, any object denoted by that label, e.g., a guitar, must have a \textit{(visual) Genus} which is described by the linguistic genus of the label and similarly for the \textit{(visual) differentia}. We often talk of linguistic and visual differentia in terms of \textit{linguistic} and \textit{visual properties}.\footnote{The distinction between the visual properties of an object and the linguistic properties which are used, in natural language, to define the label used to name the object, is based on the \emph{Teleosemantics} theory of meaning \cite{millikan2000,millikan2020neuroscience}, as adapted to CV in  \cite{2016-FOIS}.}
    
For what concerns the classifier, the goal of Choice C1 is to select the image \textit{main object} (see Section \ref{S2}). Consider for instance, in Fig.\ref{I1} (II), the objects in the MOI Image \#171, i.e., a \emph{Koto} but also a \emph{Flute}, \emph{Music Stand} etc., all labelled as \emph{Koto}. Choice C1 solves the problem of the MOI source of subjectivity. The possible strategies are: (i) removing the image just because of multi-object, (ii) splitting the image into sub-images, one per object,  or (iii) isolating objects, e.g., via bounding polygons, in this case allowing for multi-label images. Notice how Choices C1, C2 are crucially based on the distinction between \textit{object localization} and \textit{visual (image) classification}, where the former is an \textit{inherently visual activity} where an object is localized but not recognized, while the latter is an \textit{inherently semantics-driven activity} which determines the relevant properties of an object. This distinction, while being well known, is often not enforced in many state-of-the-art approaches, which collapse object detection and recognition (see \cite{he2015spatial,ILSVRC}).

Choice 2 enables the classifier to select the label whose genus and differentia align with the visual genus and differentia of the object in the image. There are only two possibilities. Either such a label does not exist, in which case the object does not belong to that class, or it does exist, in which case the opposite holds. The case of multiple possible labels cannot occur as two different objects with the same genus cannot have the same differentia. This is the key guideline
enforced by the classificationist to avoid the occurrence of the SGP many-to-many problem, modulo mistakes on the side of the annotator, i.e., the selection of a label that does not describe the contents of the image.
Thus, for instance, a positive example is Image \#238 in Fig.\ref{I1}-(II), where  the ImageNet gloss of the label \emph{Keyboard Instrument} forces the selection of the visual property of being a \emph{``a musical instrument that is played by means of a keyboard"}. Dually, Choice 2 forces the rejection of all images which do not satisfy the meaning of the label. Thus, for instance, \texttt{vTelos} will reject the MI Image \#383,  as the ImageNet label \emph{Acoustic Guitar} is defined as \emph{``a guitar with no input jack"}; but the picture does not represent a guitar and, more in general, not even a musical instrument.

\section{Annotation process}
\label{S4}
\begin{figure}[t]
\setlength{\abovecaptionskip}{7pt}%
\includegraphics[width=1\linewidth]{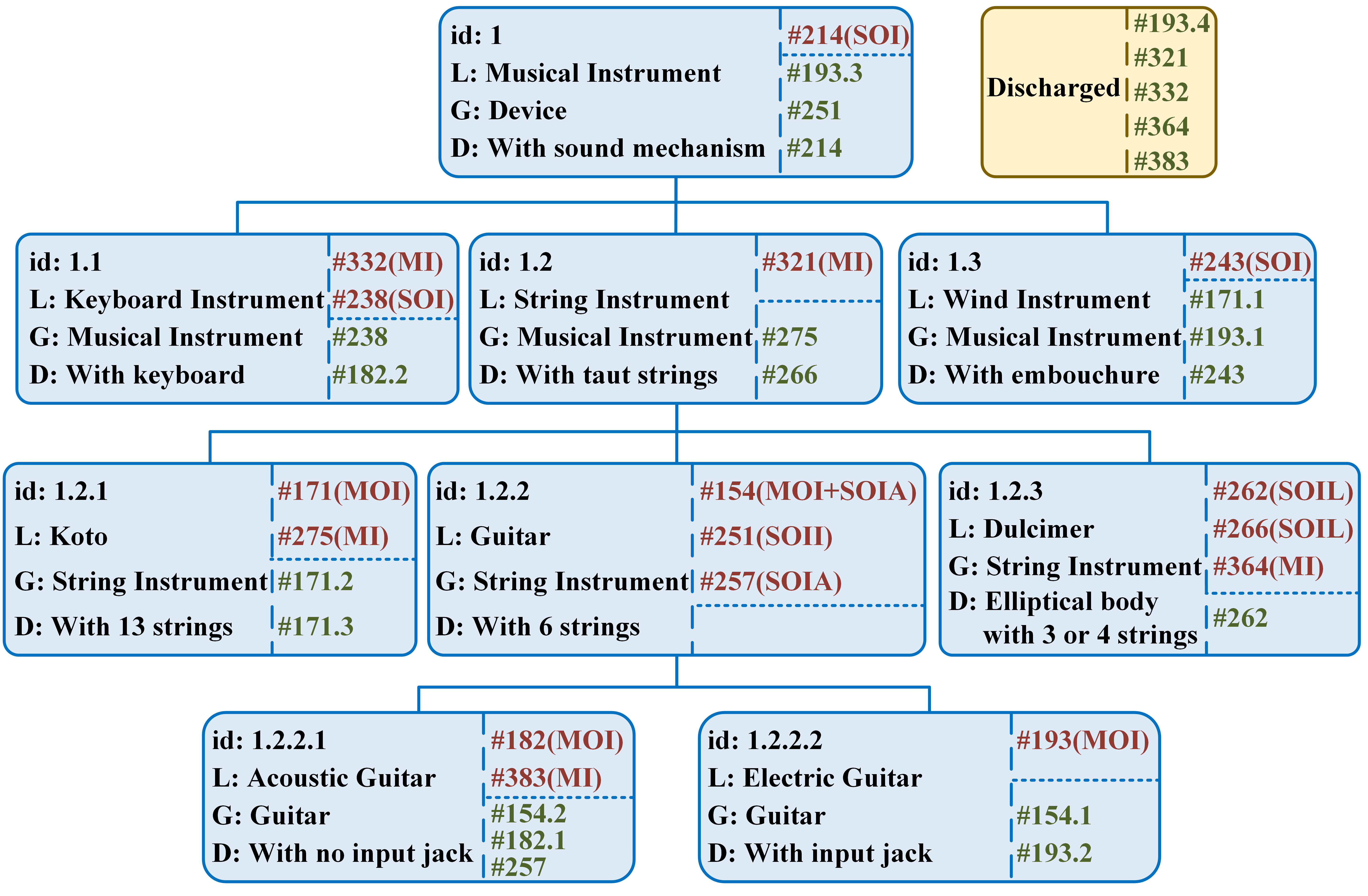}
\caption{The ImageNet \textit{Musical Instrument} hierarchy. Each node is associated with a unique linguistic identifier (id), an ImageNet label (L), a Genus (G), and a Differentia (D). Nodes are populated with the images from Fig.\ref{I1} where red and green ids correspond to the ImageNet and \texttt{vTelos} classification. The images in the yellow box are discharged by \texttt{vTelos}.}
\centering
\label{I0}
\end{figure}

\noindent
If one looks at the examples provided in Section \ref{S3}, s/he will find out that each single linguistic differentia is quite fine-grained, which is motivated by the need of making it precise and not ambiguous. S/he will also notice that, on the other hand, the linguistic genus is quite coarse-grained and ambiguous, being in fact an ImageNet label. This apparent contradiction is solved by observing that the meaning of any \textit{genus can be recursively generated in terms of a second more general genus plus multiple differentiae down to the genus itself.}  
Thus, for instance, the meaning of the genus \textit{String Instrument} of the label \textit{Koto} can be reconstructed as being \textit{a Device, with sound mechanism, with taut strings}. 
This leads to the consequence that Choices C3, C4, as carried out by the classificationist, generate a \textit{genus-differentia classification hierarchy} which suitably organizes the meaning of labels. 

The mainstream approach to the generation of this type of hierarchy is the  \textit{faceted classification} methodology \cite{SRR-67}, a tried-and-tested paradigm of developing classification schemas devised specifically to classify any type of information resource. A key innovation of this paper is the suggestion of using, as genus-differentia classification hierarchy, the \textit{WordNet} \textit{lexico-semantic hierarchy} \cite{PWN} and its evolutions, as developed in the work in Computational Linguistics. Thus, for instance, in WordNet we have the following definition of \textit{piano}:

\begin{algorithm}[tb]
    \caption{$L  = VisClassify(I)$}
    \label{alg:algorithm}
     $H$ = Tree of ($N$);\\
    $N= <id, D>$;\\    
    \textbf{Input}: Image $I$;\\
    \textbf{Output}: Image label $L$;
    \begin{algorithmic}[1] 
        \STATE $R = getRoot(H)$;
        \IF {$askDif(I, R.D) = False$}
            \RETURN $L = ``Discharged"$;    
        \ENDIF
        \STATE $L = R.D$;
            \WHILE{$hasChild(R) = True$}
                    \WHILE{$C \in getChild(R)$}
                        \IF {$askDif(I, C.D) = True$}
                        \STATE $R = C$;   $break$;                \ENDIF
                    \ENDWHILE
                \STATE $L = R.D$;   
            \ENDWHILE
            \RETURN $L$;
    \end{algorithmic}
\end{algorithm}

\begin{table}
\setlength{\abovecaptionskip}{2pt}%
\setlength{\belowcaptionskip}{0pt}%
	\resizebox{1\linewidth}{!}{%
\begin{tabular}[t]{ll}
\multicolumn{2}{c}{\textit{Word}: piano}                                                                                  \\ 
Sense ID:~03934354; piano\#1                                     & Sense ID:~04998511; piano\#2         \\ 
Synset = piano, forte-piano, ...                            & Synset = piano, pianissimo, ...  \\
Gloss = ``a keyboard instrument   & Gloss = ``(music) low loudness"   \\ 
that is played by depressing keys" &     \\ 
\end{tabular}}
\centering
\label{pwntab}
\end{table}
\vspace{-0.3cm}
\noindent
From this definition, we know that the 
 label \textit{piano} has two unique meanings, ``piano\#1"  and ``piano\#2", called \emph{senses}, each associated with a unique sense \emph{identifier}, e.g., \texttt{03934354}. Different synonymous words for the same sense are clustered into \emph{synsets}. Further, each synset is associated with a \emph{gloss} which is a natural language definition of the concept expressed by the synset in terms of genus-differentia. For instance, the synset \{\emph{piano, forte-piano, ...}\}, associated to the sense ``piano\#1", is defined by a gloss which encodes its genus \emph{``keyboard instrument"} and its differentia \emph{``depressing keys"}.
Fig.\ref{I0} reports an abstraction of the WordNet \textit{Musical Instrument} hierarchy. In turn, in ImageNet, this hierarchy has been populated with 3660 images. Notice how in Fig.\ref{I0} it is possible to reconstruct the definition of the label \textit{Koto} provided above.

The use of lexico-semantic hierarchies as genus-differentia classifications has three main consequences. The first is that 
lexico-semantic hierarchies can be reused, at no cost, for the generation of high-quality datasets (thus generalizing the idea behind the development of ImageNet).  The second is that the follow-up application of Choices C1, C2 allows to generate datasets where the semantics of language and vision are natively aligned thus enabling a general solution to the SGP many-to-many mapping problem. The third is that, given a set of images that populate the root node, \textit{the only information needed to label an image is its visual differentia}. As the following will make clear, this last fact is the one which causes a somewhat radical departure from the current mainstream approach to image annotation.

Choice C2, as carried out by the classifier, takes in input the hierarchy generated by Choices C3, C4 and associates each single image to a node in the hierarchy. The resulting algorithm is reported in Algorithm 1. This algorithm takes in input a hierarchy $H$, e.g., the WordNet/ ImageNet lexico-semantic hierarchy in Fig.\ref{I0},  
 and an Image $I$ and produces in output the image label $L$ (Lines 3, 12). After a first initialization where it checks whether the input Image can be classified, i.e., it is in the scope of the root label $R$ (as defined by differentia from a more abstract genus) (Lines 1-4), Algorithm 1 proceeds down the tree. This is achieved via a first \textit{while} (Line 6) which recursively descends the hierarchy and a second \textit{while} (Line 7) which, at each level in the hierarchy, checks whether there is a sibling matching the image visual differentia (Line 8). Algorithm 1 applies the \textit{GetSpecific} principle \cite{get-specific},
 i.e., it stops whenever none of the siblings satisfies the test in Line 8.

 Figure \ref{I0} shows the results of applying Algorithm 1 to the images in Fig \ref{I1}. Here, the images which cannot be classified in any of the nodes below the root node are collected in the yellow box and labeled \textit{Discharged}.
 Let us see how the various types of annotation mistakes from Section \ref{S2} are handled.
 We assume that MOI images have been previously split in multiple sub-images, one per object, with numbering increasing from left to right.  In fact, as from Table \ref{T1}, MOI images are dealt with by Choice C1.
We also assume that SOIL labels have been handled when building the classification hierarchy. Thus, for instance in Fig. \ref{I0},
 the label \textit{Dulcimer}, coherently with the ImageNet differentia, has a single meaning  describing \textit{American Dulcimers}. Taking into account \textit{Chinese dulcimers} would require extending ImageNet by adding a new sense to the label \textit{Dulcimer}. Synonyms are natively handled via synsets.
 MI images are moved to the label which suitably described their visual contents. Thus, for instance, Image \#364 labelled \emph{Dulcimer} in Fig.\ref{I1} (III), is discharged.
 SOII images are always classified higher in the hierarchy, if compared to the ImageNet position, this is because the partial view of the object does not allow for its full discrimination. Thus, for instance,
 Image, \#251 labelled \emph{Guitar} in Fig.\ref{I1} (II), is labelled as a \emph{musical instrument}. 
 With SOIA images the situation is the opposite of SOII images, that is, they are classified deeper in the hierarchy. Thus, for instance, Image \#257 labelled as \emph{Guitar} in Fig.\ref{I1} (II) can be correctly labelled as an \emph{Acoustic guitar} based on  the visual property \emph{``no input jack"}.
 
Based on the above, it is possible to classify the results of the annotation of an image as follows.

\begin{itemize}
    \item \textit{Correct}. The image annotation is correct;
     \item \textit{Incorrect}. The image annotation is wrong. This case can be further refined as follows:
     \begin{itemize}
         \item \textit{Generic}. The image annotation is more general than the correct label, i.e., it does not exploit all its visual properties;
          \item \textit{Restricted}. The image annotation is more specific than the correct label, i.e., it has been generated by refining the correct label using visual properties which are not visible;
           \item \textit{Misplaced}. The image annotation is unrelated to the correct label, i.e. it has been generated by using visual properties which are alternative to those needed to select the correct label.
     \end{itemize}  
\end{itemize}
\noindent
MI images can take any type of incorrect label. SOIA images always have generic labels. SOII images always have restricted labels.

The last observation is that lexico-semantic-resources are a very good starting point but more work needs to be done to make them fully usable for image annotation.  
This is in fact ongoing work. Fig.\ref{I0} exemplifies two cases. The first is the root node whose differentia is based on sound and, as such, not recognizable visually. This applies to all differentia describing senses. The second is the differentia of the label \textit{Acoustic Guitar}. This differentia, articulated as a missing visual feature, will always empty the parent node, in this case, the label \textit{guitar}.

\begin{table*}[htp]
\centering
\setlength{\abovecaptionskip}{2pt}%
\setlength{\belowcaptionskip}{0pt}%
	\resizebox{1\linewidth}{!}{%
\begin{tabular}{|cccccccccccc|cccccccccc|}
\hline
\multicolumn{1}{|c|}{\multirow{2}{*}{\textbf{Id.}}}  & \multicolumn{1}{c|}{\multirow{2}{*}{\textbf{GT1}}} & \multicolumn{1}{c|}{\multirow{2}{*}{\textbf{GT2}}}  &  \multicolumn{9}{c|}{\textbf{GT3 (Annotation via Differentia labels)}}                                                                                                                                      & \multicolumn{9}{c|}{\textbf{GT4 (Annotation via Category labels)}}       \\ \cline{4-21}
\multicolumn{1}{|c|}{}            & \multicolumn{1}{c|}{}                    & \multicolumn{1}{c|}{}      &    \multicolumn{1}{c|}{\textbf{Differentia}}                          & \textbf{NA$_{1.1}$}  & \textbf{NA$_{1.2}$}  & \textbf{NA$_{1.3}$}  & \textbf{NA$_{1.4}$}  & \textbf{NA$_{1.5}$}  & \textbf{NA$_{1.6}$}  & \textbf{NA$_{1.7}$}  & \multicolumn{1}{c|}{\textbf{NA$_{1.8}$} } &      \multicolumn{1}{c|}{\textbf{Categories}}          & \textbf{NA$_{2.1}$} & \textbf{NA$_{2.2}$} & \textbf{NA$_{2.3}$} & \textbf{NA$_{2.4}$} & \textbf{NA$_{2.5}$} & \textbf{NA$_{2.6}$} & \textbf{NA$_{2.7}$} & \multicolumn{1}{c|}{\textbf{NA$_{2.8}$}}       \\ \hline
\multicolumn{1}{|l|}{1}       & \multicolumn{1}{c|}{50}    & \multicolumn{1}{c|}{41}   &    \multicolumn{1}{|l|}{with Sound Mechanism}    & 33          & 12          & 27          & 25          & 28          & 29          & 12          & \multicolumn{1}{c|}{18}                    & \multicolumn{1}{l|}{Musical Instrument}  & 16          & 42          & 100         & 17          & 27          & 20          & 19          & \multicolumn{1}{c|}{26}     \\
\multicolumn{1}{|l|}{1\_1}      & \multicolumn{1}{c|}{50}      & \multicolumn{1}{c|}{123}     &  \multicolumn{1}{|l|}{with Taut Strings}                   & 46          & 97          & 71          & 133         & 112         & 83          & 62          & \multicolumn{1}{c|}{79}         & \multicolumn{1}{l|}{Stringed Instrument} & 162         & 78          & 0           & 115         & 144         & 106         & 161         & \multicolumn{1}{c|}{74}      \\
\multicolumn{1}{|l|}{1\_1\_1}      & \multicolumn{1}{c|}{50}      & \multicolumn{1}{c|}{34}      &  \multicolumn{1}{|l|}{with 6 Strings}                       & 37          & 13           & 40          & 34          & 58          & 34          & 19           & \multicolumn{1}{c|}{31}         & \multicolumn{1}{l|}{Guitar}              & 77          & 18          & 41          & 40          & 13          & 9           & 0           & \multicolumn{1}{c|}{8}       \\
\multicolumn{1}{|l|}{1\_1\_1\_1}     & \multicolumn{1}{c|}{50}      & \multicolumn{1}{c|}{40}         &  \multicolumn{1}{|l|}{with No Input Jack}                     & 66          & 77        & 53          & 54          & 26          & 67        & 68          & \multicolumn{1}{c|}{50}       & \multicolumn{1}{l|}{Acoustic Guitar}     & 13          & 81          & 66          & 43          & 70          & 70          & 82          & \multicolumn{1}{c|}{71}       \\
\multicolumn{1}{|l|}{1\_1\_1\_2}      & \multicolumn{1}{c|}{50}     & \multicolumn{1}{c|}{43}     & \multicolumn{1}{|l|}{with Input Jack}                          & 54          & 61          & 67          & 43          & 28          & 45         & 82          & \multicolumn{1}{c|}{78}       & \multicolumn{1}{l|}{Electric Guitar}     & 74          & 65          & 86          & 44          & 70          & 71          & 68          & \multicolumn{1}{c|}{74}    \\
\multicolumn{1}{|l|}{1\_1\_2}     & \multicolumn{1}{c|}{50}      & \multicolumn{1}{c|}{31}    & \multicolumn{1}{|l|}{Elliptical body with 3 or 4 strings} & 61          & 37          & 36          & 32          & 29          & 30         & 37          & \multicolumn{1}{c|}{30}          & \multicolumn{1}{l|}{Dulcimer}            & 0           & 6           & 31          & 31          & 0           & 16          & 6           & \multicolumn{1}{c|}{46}     \\
\multicolumn{1}{|l|}{1\_1\_3}      & \multicolumn{1}{c|}{50}      & \multicolumn{1}{c|}{27}   & \multicolumn{1}{|l|}{with 13 Strings}                         & 42          & 32          & 42          & 16          & 18          & 44          & 53          & \multicolumn{1}{c|}{45}           & \multicolumn{1}{l|}{Koto}                & 0           & 47          & 47          & 39          & 0           & 41          & 0           & \multicolumn{1}{c|}{43}          \\
\multicolumn{1}{|l|}{1\_2}      & \multicolumn{1}{c|}{50}     & \multicolumn{1}{c|}{47}      & \multicolumn{1}{|l|}{with Keyboard}                      & 49          & 46          & 41          & 49          & 43          & 47          & 50          & \multicolumn{1}{c|}{46}       & \multicolumn{1}{l|}{Keyboard Instrument} & 41          & 49          & 0           & 44          & 40          & 47          & 52          & \multicolumn{1}{c|}{44}       \\
\multicolumn{1}{|l|}{1\_3}         & \multicolumn{1}{c|}{50}     & \multicolumn{1}{c|}{47}    & \multicolumn{1}{|l|}{with Embouchure}             & 62          & 60          & 60          & 63          & 50          & 55          & 60          & \multicolumn{1}{c|}{59}             & \multicolumn{1}{l|}{Wind Instrument}     & 61          & 57          & 55          & 62          & 54          & 61          & 60          & \multicolumn{1}{c|}{59}      \\
\multicolumn{1}{|l|}{}     & \multicolumn{1}{c|}{0}  & \multicolumn{1}{c|}{17}       & \multicolumn{1}{|l|}{Unrecognised}                         & 0           & 15          & 13          & 1           & 58          & 12          & 7           & \multicolumn{1}{c|}{14}          & \multicolumn{1}{l|}{Unrecognised}        & 6           & 7           & 24          & 15          & 32          & 9           & 2           & \multicolumn{1}{c|}{5}    \\ \hline
\multicolumn{1}{|l|}{all}   &   \multicolumn{1}{c|}{450}              & \multicolumn{1}{c|}{450}     & \multicolumn{1}{|l|}{Krippendorff’s alpha}          & \multicolumn{8}{|c|}{0.5973}       & \multicolumn{1}{l|}{Krippendorff’s alpha}                                                                                                       & \multicolumn{8}{c|}{0.5047}      \\ \hline
\end{tabular}
}
\vspace{0.2cm}
\caption{Inter-annotator agreement evaluation with Choice C2 only. The images in GT1 have ImageNet labels. GT2, GT3, GT4 images have labels selected by the expert annotator, and by the 8+8 non-expert annotators in Groups 1 and 2, respectively. NA$_{j,i}$ stands for annotator $i$ of Group $j$. Id. is the label identifier, as from Fig. 2. The last row reports the value of Krippendorff's alpha.}
\label{tab:3}
\end{table*}

\begin{table}
\centering
\setlength{\abovecaptionskip}{2pt}%
\setlength{\belowcaptionskip}{0pt}%
	\resizebox{1\linewidth}{!}{%
\begin{tabular}{|l|l|cc|cccccccc|}
\hline
\multicolumn{1}{|c|}{\multirow{2}{*}{\textbf{Index}}} &  \multicolumn{1}{|c|}{\multirow{2}{*}{\textbf{Categories}}}
& \multicolumn{2}{|c|} {\textbf{GT2*}}    
& \multicolumn{8}{c|}{\textbf{GT3* (Only Single-Object images)}}     \\ \cline{3-12} 
\multicolumn{1}{|c|}{}      &       & \textbf{EA$_1$}& \textbf{EA$_2$}    & \textbf{NA$_{1.1}$} & \textbf{NA$_{1.2}$} & \textbf{NA$_{1.3}$} & \textbf{NA$_{1.4}$} & \textbf{NA$_{1.5}$} & \textbf{NA$_{1.6}$} & \textbf{NA$_{1.7}$} & \multicolumn{1}{c|}{\textbf{NA$_{1.8}$}}\\ \hline

1                                                                                & \multicolumn{1}{l|}{with   Sound Mechanism}      & 17   &  17               & 16            & 2             & 5             & 11            & 11            & 11            & 4             & \multicolumn{1}{c|}{4}             \\
1\_1                                                                              & \multicolumn{1}{l|}{with Taut   Strings}           & 42 &42             & 21            & 41            & 32            & 43            & 41            & 39            & 28            & \multicolumn{1}{c|}{32}         \\
1\_1\_1                                                                           & \multicolumn{1}{l|}{with 6 Strings}                & 21 & 20            & 21            & 8             & 23            & 19            & 26            & 24            & 11            & \multicolumn{1}{c|}{15}             \\
1\_1\_1\_1                                                                        & \multicolumn{1}{l|}{with No Input   Jack}       & 21  & 22               & 31            & 34            & 29            & 28            & 18            & 32           & 33            & \multicolumn{1}{c|}{29}            \\
1\_1\_1\_2                                                                       & \multicolumn{1}{l|}{with Input Jack}                 & 22   &  22            & 24            & 32            & 31            & 21            & 20            & 21            & 39            & \multicolumn{1}{c|}{39}        \\
1\_1\_2                                                                           & \multicolumn{1}{l|}{Elliptical body with 3 or 4 strings}         & 13   &  13             & 24            & 13            & 13            & 14            & 12            & 13          & 15            & \multicolumn{1}{c|}{14}            \\
1\_1\_3                                                                           & \multicolumn{1}{l|}{with 13 Strings}             & 12  &  12                & 12            & 7             & 10            & 8             & 9             & 10            & 13            & \multicolumn{1}{c|}{9}            \\
1\_2                                                                              & \multicolumn{1}{l|}{with Keyboard}                  & 33  &  33               & 33            & 33            & 29            & 34            & 29            & 31            & 34            & \multicolumn{1}{c|}{33}       \\
1\_3                                                                              & \multicolumn{1}{l|}{with Embouchure}             & 10    &  10                & 20            & 20            & 21            & 23            & 14            & 15           & 19            & \multicolumn{1}{c|}{21}           \\
 & \multicolumn{1}{l|}{Unrecognised}                 & 11   &  11             & 0             & 12            & 9             & 1             & 22            & 6             & 6             & \multicolumn{1}{c|}{6}            \\ \hline
all    &   \multicolumn{1}{l|}{Krippendorff’s alpha}                                                                                                                & \multicolumn{2}{c|}{0.9877}                                             & \multicolumn{8}{c|}{0.7595}        \\ \hline
\end{tabular}}
\vspace{0.2cm}
\caption{Inter-annotator agreement with Choices C1, C2 (full \texttt{vTelos}), generating two GT2* and eight GT3* datasets.}
\label{tab:6}
\end{table}

\begin{table}
\centering
\setlength{\abovecaptionskip}{2pt}%
\setlength{\belowcaptionskip}{3pt}%
	\resizebox{1\linewidth}{!}{%
\begin{tabular}{|c|l|l|l|}
\hline
\textbf{Anno.} & \textbf{Acoustic   Guitar}        & \textbf{Dulcimer}                                   & \textbf{Koto}                          \\ \hline
\textbf{NA$_{1.1}$}          & Guitar                            & Appalachian   Dulcimer                              & Biwa                                   \\
\textbf{NA$_{1.2}$}          & Guitar                            & IDK (I Don't Know)                                                 & IDK                                    \\
\textbf{NA$_{1.3}$}          & Wooden   guitar                   & IDK                                                 & Zither                                 \\
\textbf{NA$_{1.4}$}         & Wooden   guitar                   & 3 or 4 String Musical Instrument             & 13 String Koto                 \\
\textbf{NA$_{1.5}$}         & Hawaiian   Guitar                 & 3-4 Stringed Elliptical Instrument   & 13-String Instrument \\
\textbf{NA$_{1.6}$}          & 6 Stringed Instrument no Jack     & Fretted Stringed Instrument                       & 13 Stringed Instrument         \\
\textbf{NA$_{1.7}$}          & Classic   Guitar                  & Elliptical Stringed Instrument                    & Rectangular   Stringed Instrument      \\
\textbf{NA$_{1.8}$}         & Non   Powered Guitars             & Short-Stringed Music Instruments                    & Japanese   Stringed Instrument        \\ \hline
\end{tabular}}
\vspace{0.2cm}
\caption{Category Labels suggested by Group 1 annotators.}
\label{tab:4}
\end{table}

\section{Evaluation}
\label{S5}

As from Section \ref{I1}, all the experiments focus on ImageNet.   We focus only on the annotator choices C1, C2, where choices C3, C4 are automatically enforced by the fact that we use ImageNet as background knowledge. To this extent, in Choice C2, we assume the visual differentia to be as defined by the Imagenet gloss.
We perform four evaluations.  In the first, we evaluate  the inter-annotator agreement, as a proxy of the decrease of the role of subjective choices. We consider both expert and non-expert annotators. In the second we evaluate the annotation cost.
In the third, we evaluate the increase of accuracy generated on various state-of-the-art ML model. The last experiment is an ablation study.

\subsection{Inter-annotator agreement}
\label{label.2}

We selected 450 of the 3660 images populating the ImageNet categories in Fig.\ref{I0}, 50 images per category. To minimize data bias, these 450 labelled image dataset, let us call GT1 (where GT stands for \textit{Ground Truth}) was randomly generated. 
Then, a second dataset GT2 was generated by an expert annotator EA$_1$ by relabelling the 450 images of GT1 using  Choice C2 only, as described in Section \ref{S4}. We then selected 16 volunteer \textit{non-expert annotators}, with no previous annotation experience. The annotators were university students and collaborators, with an average age of 29 years. To maximize diversity, we selected people from different disciplines and from 3 continents and 7 countries. The annotators were organised in 2 groups of 8 people, each annotator relabelling GT1 and generating a dataset as follows:

\begin{itemize}
\item GT3, as created by each member of Group 1 The annotators were 
asked to perform the annotation in two steps. First, they had to annotate images based on Choice C2.  Here they were allowed to create a new category ``\textit{Discharged}" for all the images they could not classify. Then, in the second step, they were asked to label the categories generated in the first step S2 with their ``most suitable" label, not necessarily an ImageNet label (which was not provided).

\item GT4, as created by each member of Group 2. This group was instructed to annotate images using the ImageNet labels plus the label ``Discharged".
\end{itemize}
The motivation for focusing EA$_1$ and Group 1 on Choice C2 was that of highlighting the use of the differentia in place of the label.
The results are reported in Table \ref{tab:3}. Notice how this table reports the number of images associated with each label for all four datasets and for each annotator working on GT3 and GT4. As it can be seen, the number of images (and the images) associated with the same label are different. We measure the inter-annotator agreement of GT3 and GT4 using \textit{Krippendorff's alpha}  \cite{k-alpha,artstein2017inter} (last row).
 Krippendorff's alpha is $0.5973$ for GT3 and $0.5047$ for GT4, that is, GT3 is around 18\% higher than GT4. This is an important improvement but still not good enough.\footnote{As from \cite{hughes2021krippendorffsalpha}[Table 2, p. 417], a value of alpha in the range $(0.6, 0.8]$  indicates substantial agreement while an index in the range $(0.8, 1]$  indicates near-perfect agreement.} The cause of this, not fully satisfactory, result was the Multi-object images which are more than 58\% of the total (2125 out of 3660 images). 
We have therefore performed a second experiment where Choice C1, C2 were enforced and where MOI images were proposed multiple times with a bounding box around the object to be annotated. This experiment was performed by Group 1 (generating eight datasets GT2*) and two expert annotators EA$_1$, EA$_2$ (generating two datasets GT3*).
Table \ref{tab:6} reports the results obtained. Here, Krippendorff's alpha goes up to $0.7595$ among non-expert annotators and $0.9877$ between the two experts, namely up to near-perfect agreement. 

An interesting  final observation is that in Group 2, but not in Group 1, 
some annotators were unable to annotate some images (see, e.g., the ``0"s in Table \ref{tab:3}). Two such examples are   ``Dulcimer" and ``Koto".\footnote{Koto is a Japanese instrument.}
This observation is also confirmed in Table \ref{tab:4}, which reports some example labels provided during the second part of the Group 1 experiment. The observation here is that, despite properly labeling images via differentia, some annotators did not know the names of some categories, or used wrong or more generic labels, or in one case used a more specific label.

\begin{table*}[h]
\centering
\setlength{\abovecaptionskip}{0pt}%
\setlength{\belowcaptionskip}{0pt}%
	\resizebox{1\linewidth}{!}{%
\begin{tabular}{|l|cccccccccccccccc|}
\hline
\multirow{2}{*}{\textbf{Methods}} & \multicolumn{2}{c|}{\textbf{AlexNet}} & \multicolumn{2}{c|}{\textbf{ZFNet}} & \multicolumn{2}{c|}{\textbf{VGG}}      & \multicolumn{2}{c|}{\textbf{GoogleNet}} & \multicolumn{2}{c|}{\textbf{ResNet}}   & \multicolumn{2}{c|}{\textbf{DenseNet}} & \multicolumn{2}{c|}{\textbf{RAN}}     & \multicolumn{2}{c|}{\textbf{SENets}} 
\\ \cline{2-17} 
  & \textbf{Acc.} & \multicolumn{1}{c|}{\textbf{Impro.(\%)}}  & \textbf{Acc.} & \multicolumn{1}{c|}{\textbf{Impro.(\%)}} & \textbf{Acc.} & \multicolumn{1}{c|}{\textbf{Impro.(\%)}}     & \textbf{Acc.} & \multicolumn{1}{c|}{\textbf{Impro(\%)}} & \textbf{Acc.} & \multicolumn{1}{c|}{\textbf{Impro.(\%)}}   & \textbf{Acc.} & \multicolumn{1}{c|}{\textbf{Impro.(\%)}} & \textbf{Acc.} & \multicolumn{1}{c|}{\textbf{Impro.(\%)}}   & \textbf{Acc.} & \multicolumn{1}{c|}{\textbf{Impro.(\%)}} \\ \hline
                                  
\textbf{ImageNet}                 & 0.543   & \multicolumn{1}{c|}{-}  & 0.612  & \multicolumn{1}{c|}{-} & 0.655    & \multicolumn{1}{c|}{-}  & 0.734     & \multicolumn{1}{c|}{-}  & 0.593    & \multicolumn{1}{c|}{-}  & 0.724   & \multicolumn{1}{c|}{-}   & 0.713    & \multicolumn{1}{c|}{-} & 0.728              & -           \\
\textbf{only Chioce C1}     & 0.551   & \multicolumn{1}{c|}{1.47\%}  & 0.623  & \multicolumn{1}{c|}{1.80\%} & 0.672 & \multicolumn{1}{c|}{2.55\%}  & 0.745  & \multicolumn{1}{c|}{1.55\%}  & 0.610 & \multicolumn{1}{c|}{2.86\%}  & 0.733   & \multicolumn{1}{c|}{1.24\%}   & 0.718  & \multicolumn{1}{c|}{0.74\%} & 0.746          & 2.54\%           \\
\textbf{only Chioce C2}       & 0.583   & \multicolumn{1}{c|}{7.37\%}  & 0.644  & \multicolumn{1}{c|}{5.23\%} & 0.732 & \multicolumn{1}{c|}{11.82\%} & 0.829     & \multicolumn{1}{c|}{12.94\%} & 0.729    & \multicolumn{1}{c|}{22.93\%} & 0.782   & \multicolumn{1}{c|}{8.01\%}   & 0.770 & \multicolumn{1}{c|}{8.04\%} & 0.807           & 10.87\%          \\
\textbf{Choice C1 \& C2 (vTelos)}                   & 0.596   & \multicolumn{1}{c|}{9.76\%}  & 0.657  & \multicolumn{1}{c|}{7.35\%} & 0.743    & \multicolumn{1}{c|}{13.44\%} & 0.835     & \multicolumn{1}{c|}{13.76\%} & 0.732    & \multicolumn{1}{c|}{23.44\%} & 0.793   & \multicolumn{1}{c|}{9.53\%}   & 0.784    & \multicolumn{1}{c|}{9.96\%} & 0.811              & 11.40\%          \\ \hline
\end{tabular}}
\vspace{0.2cm}
\caption{Ablation study. ``Acc."  is the accuracy, ``Impro.(\%)"  is the improvement in performance over ImageNet.} 
\label{tab:7}
\end{table*}

\begin{table}
\centering
\setlength{\abovecaptionskip}{2pt}%
\setlength{\belowcaptionskip}{3pt}%
\resizebox{1\linewidth}{!}{%
\begin{tabular}{|l|cc|c|}
\hline
\multirow{2}{*}{\textbf{Methods}} & \multicolumn{2}{c|}{\textbf{Accuracy}}     &   \multirow{2}{*}{\textbf{Improvement}}            \\ \cline{2-3} 
                                 & \textbf{ImageNet} & \multicolumn{1}{c|}{\textbf{vTelos}}  & \\ \hline
AlexNet \cite{alexnet}           & 0.543             & 0.596            & 9.76\%   \\
ZFNet \cite{zfnet}               & 0.612             & 0.657            & 7.35\%   \\
VGG \cite{vgg}                   & 0.655             & 0.743            & 13.44\%   \\
GoogleNet \cite{googlenet}       & 0.734             & 0.835            & 13.76\%   \\
ResNet \cite{resnet}             & 0.593             & 0.732            & 23.44\%  \\
DenseNet \cite{densenet}         & 0.724             & 0.793            & 9.53\%  \\
RAN \cite{ran}                   & 0.713             & 0.784            & 9.96\%   \\
SENets \cite{senet}              & 0.728             & 0.811            & 11.40\%   \\ \hline
\end{tabular}}
\vspace{0.2cm}
\caption{Classification results for ImageNet and  \texttt{vTelos}.}
\label{tab:5}
\end{table}

\subsection{Annotation cost}

As from Section \ref{S4}, \texttt{vTelos} requires that each image is recursively compared with the node one level down in the lexico-semantic hierarchy till it meets a differentia which it does not satisfy. So the question arises of what is the cost of the multiple checks. The preliminary observation is that, while containing an excess of 100K concepts, WordNet is a broad and flat hierarchy whose paths have a depth in the range of around 7 to 15 nodes (with different depths for different versions) and where the nodes higher in the hierarchy are very abstract and have no visual content. 
A 4-node deep hierarchy, like the one in Fig. \ref{I0}, is therefore a good proxy for evaluating the annotation cost.
We have therefore compared the time consumed by two non-expert annotators in the generation of two GT3 and GT4-like datasets restricted to contain one hundred images. The results show that the average annotation time per image is 5.565 seconds for GT3 and 3.473 seconds for GT4, that is, a time increase for \texttt{vTelos} of around 60\% which one would guess it would become a little more than double with 8-node level deep hierarchies. The time increase correlates to the images classified in intermediate nodes. The fact that it is a low increase seems to correlate to the fact that each single choice is simpler given that: (i) the same image is seen more than once, (ii) the choice is restricted within a precise set of labels, (iii) which have similar meanings, because selected following the pattern induced by the tree structure, thus ultimately improving the automation of the process. 

\subsection{Machine Learning Accuracy}
\label{label.3}

We applied eight state-of-the-art ML methods trained on two datasets, called below ``ImageNet"  ``\texttt{vTelos}, the first containing the original  3660 images, the second containing the same 4605 
images where MOI images are transformed in SOI images. 80\% of the images were used as training set and the remaining 20\% were the test set.

All the experiments have been implemented with PyTorch with identical settings. During the  training,  we performed the same data augmentation (horizontal flipping and random scaling with multiple times) on each dataset and randomly sampled $224 \times 224$ crops from augmented images. All experiments have been optimized using Adaptive Moment Estimation 
with learning rate 
initialized to $0.0002$, momentum initialized 
to $0.9$, and weight decay 
to $10^{-8}$. All models have been pre-trained on ImageNet with ResNet101 \cite{resnet}. 
The results are reported in Table~\ref{tab:5}. It can be noticed that all models trained on \texttt{vTelos} get significant improvement in accuracy, especially ResNet which achieves up to 23.44\% improvement.

Finally, if we look at the heat maps in 
Fig.~\ref{heat-map}, we can see how the models  pay attention to the visual regions described by the linguistic differentia, see for instance the keyboard in the case of the label \textit{Keyboard Instrument}.

\subsection{Ablation study}

\begin{figure}[t]
\centering
\includegraphics[width=1\linewidth]{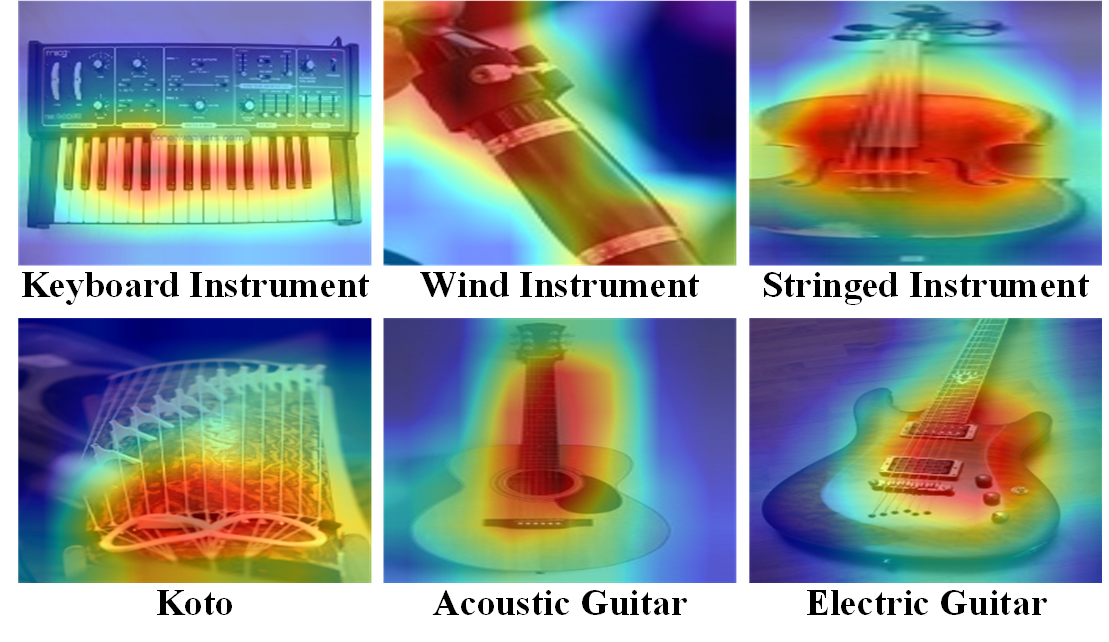}
\caption{Attention heat maps.}
\label{heat-map}
\end{figure}

We conducted two ablation studies, the first focusing on Choice C1, the second of Choice C2 for all the eight systems tested in the previous experiment. The results are reported in Table \ref{tab:7} where the first and the last row replicate the results in Table \ref{tab:5}.
The key observation is that, in all cases, the improvement in performance due to Choice C2 with respect to that due to Choice C1 is substantial (e.g., 5 times with AlexNet, 2.9 times with ZFNet, 13.5 times with RAN, 8 times with ResNet) thus providing further confirmation of the intuition suggested by the heat maps in Fig. \ref{heat-map}. Taking into account  the relevance of Choice C1 which transforms Multi-Object Images into Single-Object images, we take this as an important indicator of the fact that \textit{labeling images using differentia labels is the way to go.}

\section{Related work}
\label{S6}
In this work, we build on top of the ImageNet work~\cite{IMAGENET-2009}. With respect to this work, the main innovations are as follows. First, the idea of using (differentia) properties, and not class names, for labeling images. Second, the idea of defining an annotation methodology, rather than just producing a data set. Third, exploiting the faceted classification approach as a powerful technique for further improving the annotation quality. Finally, the fact that our approach allows for the generation of natural language aware annotations. As a consequence, the relabeling of the ImageNet images will open the door to the possibility of a dataset labeled in up to 1000+ languages, including a large number of minority languages. 

As from the introduction, the problem of dataset quality has been known for a while, with \cite{torralba2011unbiased} being an early paper focusing on it and \cite{SGP-2000} being the work which identified the SGP as the main cause of the problem. Recent work has started focusing on ways for dealing with the problem of dataset quality. Thus, \cite{2021-MLDatasetDev} advocates more careful practices in the development of datasets that are attentive to their limitations and impact. \cite{2020-ACMFAT} advocates for fairer datasets and provides a vertical methodology for balancing the people subtree of ImageNet. \cite{yun2021re} shows how to improve the quality of ImageNet by using a classifier trained in a higher-quality dataset; this work is complementary to ours as it leaves open the problem of how to generate the higher quality dataset. As far as we know, \texttt{vTelos} is the first methodology for generating high-quality datasets and it does so by using a highly inter-disciplinary KR/CV/NLP approach.  

With respect to the annotation process, the crowdsourcing community has focused extensively on the problem of quality, see, e.g., \cite{nowak2010reliable,ewerth2017machines}.  \cite{daniel2018quality} provides a quite comprehensive characterization of quality in crowdsourcing and an extensive analysis of the state of the art. Very recently some work has started to focus on how to operationally improve the quality of the process, see, e.g.,  \cite{kyriakou2021crowdsourcing,demartini2021managing}. Most relevant to this paper, \cite{nassar2019assessing} provides and exploits a set of metrics, including Krippendorff's alpha, with the goal of monitoring the image annotation process.
However, as far as we know, there two are key differences. The first is that \texttt{vTelos} is an end-to-end general methodology. The second is that, differently from the work above, which focuses on how to measure the effects and control the behaviour of annotators, here the focus is on aligning the semantics encoded in images and in NLP descriptions, i.e., in dealing with the SGP many-to-many mappings.

Earlier work on the SGP has focused on how to integrate feature-level and semantic-level information. Thus, some have proposed the use of ontologies \cite{hare2006mind}, others of high-level features \cite{ma2010bridging,elahi2017exploring}, and others to ask users \cite{tang2011semantic}. More recently  \cite{pang2019} has proposed to handle the SGP when aggregating multi-level features. All these approaches focus on object labels and not on the differentia. 

From a methodological point of view, this work is based on a mix of KR and NLP. In KR, as applied to CV, a fair amount of work, motivated by (Cognitive) Robotic applications has concentrated on identifying the function of objects see, e.g., 
\cite{bogoni1995interactive,pechuk2005function,levesque2008cognitive}. 
None of this work uses the faceted classification approach. Furthermore, a second difference is that this work has concentrated on how to enable a meaningful human-machine interaction and not, as it is the case here, on how to enforce coherence between how humans and machines describe and name objects. A lot of work has also focused on the integration of NLP and CV, see \cite{CV-2021-mogadala} for a recent very extensive survey of the field. While none of this work is focused on the definition of a general methodology for high quality dataset generation, image captioning could be used towards the automatic generation of glosses. This would be relevant when there is a need for new corpora for which human annotation is not an option (for instance because requiring expert annotators).

\section{Conclusion}
\label{S7}

In this paper, we have proposed a general CV/NLP/KR  methodology for producing high quality image datasets. The motivation for this work lies in the need of overcoming some of the limitations which exist in the current annotation process. 
This is only a first step. We believe in fact that the use of NLP/KR techniques, as the main mechanism for labeling examples, is the key to the development of explainable and high performance ML systems. 

\section*{Acknowledgements}
This research has received funding from the European Union's Horizon 2020 FET Proactive project ``WeNet – The Internet of us”, grant agreement No 823783.


\bibliography{ecai}
\end{document}